\documentclass[10pt,journal]{IEEEtran}

\usepackage[pdftex]{graphicx}
\usepackage{algorithm}
\usepackage{color}
\usepackage{caption}
\usepackage{algpseudocode}
\definecolor{ForestGreen}{RGB}{162,52,0}

\usepackage{xcolor}
\usepackage{amsmath}
\usepackage{amssymb}
\usepackage{latexsym}
\usepackage{CJK}
\usepackage{url}
\usepackage{array}
\usepackage{ragged2e}
\usepackage{makecell}
\usepackage{multirow}
\usepackage{threeparttable}

\ifCLASSOPTIONcompsoc
  \usepackage[nocompress]{cite}
\else
  \usepackage{cite}
\fi

\ifCLASSINFOpdf
\else
\fi

\ifCLASSOPTIONcompsoc
  \usepackage[caption=false,font=footnotesize,labelfont=sf,textfont=sf]{subfig}
\else
  \usepackage[caption=false,font=footnotesize]{subfig}
\fi

\begin{document}

\title{Deep Reinforcement Learning Assisted Federated Learning Algorithm for Data Management of IIoT}

\author{Peiying Zhang, Chao Wang, Chunxiao Jiang, and Zhu Han
%,~\IEEEmembership{Senior Member,~IEEE}

\thanks{This work is partially supported by the National Key Research and Development Program of China under Grant 2020YFB1804800, partially supported by the Major Scientific and Technological Projects of CNPC under Grant ZD2019-183-006, and partially supported by Shandong Provincial Natural Science Foundation under Grant ZR2020MF006. \textit{(Corresponding authors: Peiying Zhang and Chunxiao Jiang)}.}
\thanks{Peiying Zhang and Chao Wang are with the College of Computer Science and Technology, China University of Petroleum (East China), Qingdao 266580, China. E-mail: zhangpeiying@upc.edu.cn and wangch\_upc@qq.com.}
\thanks{Chunxiao Jiang is with the School of Information Science and Technology, Tsinghua University, Beijing 100084, China. E-mail: jchx@tsinghua.edu.cn.}
\thanks{Zhu Han is with the Department of Electrical and Computer Engineering, University of Houston, USA. E-mail: zhan2@uh.edu.}
}

\markboth{IEEE Transactions on Industrial Informatics,~Vol.~XX, No.~XX, XX~2020}
{}

\maketitle
\begin{abstract}
The continuous expanded scale of the industrial Internet of Things (IIoT) leads to IIoT equipments generating massive amounts of user data every moment. According to the different requirement of end users, these data usually have high heterogeneity and privacy, while most of users are reluctant to expose them to the public view. How to manage these time series data in an efficient and safe way in the field of IIoT is still an open issue, such that it has attracted extensive attention from academia and industry. As a new machine learning (ML) paradigm, federated learning (FL) has great advantages in training heterogeneous and private data. This paper studies the FL technology applications to manage IIoT equipment data in wireless network environments. In order to increase the model aggregation rate and reduce communication costs, we apply deep reinforcement learning (DRL) to IIoT equipment selection process, specifically to select those IIoT equipment nodes with accurate models. Therefore, we propose a FL algorithm assisted by DRL, which can take into account the privacy and efficiency of data training of IIoT equipment. By analyzing the data characteristics of IIoT equipments, we use MNIST, fashion MNIST and CIFAR-10 data sets to represent the data generated by IIoT. During the experiment, we employ the deep neural network (DNN) model to train the data, and experimental results show that the accuracy can reach more than 97\%, which corroborates the effectiveness of the proposed algorithm.
\end{abstract}


\IEEEpeerreviewmaketitle

\section{Introduction}

\begin{figure*}[!h]
\centering
\includegraphics[width=0.85\textwidth]{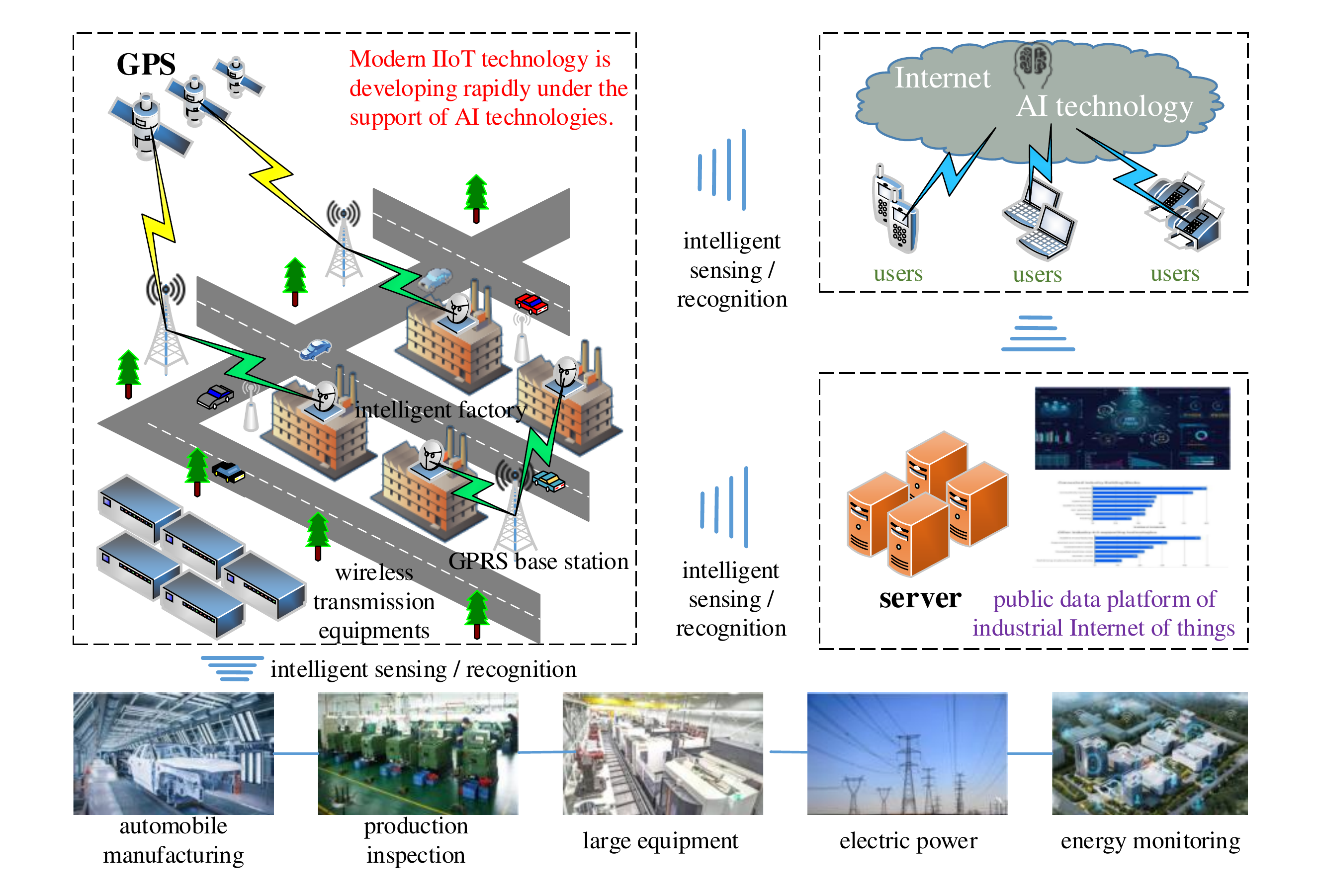}
\caption{Industrial Internet of Things scene in the context of modern society.}
\label{fig_1}
\end{figure*}

Human society is rapidly moving towards the era of industry 4.0 \cite{1}. The global distribution of user equipments (UEs) is widespread and decentralized due to the influence of geographic location. Limited by the costs of transmission media (cables, optical fibers) and communication delays, traditional network infrastructures are not suitable for the development of Industry 4.0 \cite{2}. On the other hand, wireless networks are widely used in Industry 4.0 because of its flexibility and portability. The industrial Internet of Things (IIoT) is one key technology to realize Industry 4.0. Many applications of IIoT are based on wireless networks, such as intelligent robots, driverless cars, smart grid and smart medical \cite{two4}. Fig. \ref{fig_1} shows the IIoT scenario under the background of rapid development of social science and technology. A large number of IIoT equipments access to IIoT frequently, which can produce a huge amount of data in a short period of time \cite{3,two3}. Radio network resource management faces severe challenges, including storage, spectrum, computing resource allocation, and joint allocation of multiple resources \cite{jcx1,jcx2}. With the rapid development of communication networks, the integrated space-ground network has also become a key research object \cite{jcx3}. How to efficiently manage, store and use these time series data has become an important research topic.

The data generated by IIoT equipments often involves users' private information \cite{j1,a3}. Service providers and users do not want this information to be exposed to the third party, but this kind of data is usually vulnerable to attacks from heterogeneous networks, heterogeneous equipments or malware \cite{zz1}. Therefore, there needs an approach to support IIoT, that cannot only maintain the heterogeneity and privacy of data, but also reduce the communication cost and model training deviation \cite{5,z3}. As a new type of machine learning (ML) paradigm, federated learning (FL) has attracted great attention in the industry \cite{6}. FL prevents the leakage of users' personal private information to a certain extent by separating the central server's direct access to the original data from the model training \cite{7,8}. In addition, FL can effectively maintain the heterogeneity of data and reduce the deviation of model training. Reference \cite{two1} has proved that FL can be effectively applied to multiple learning tasks. Considering the privacy, heterogeneity and wide distribution of the device data in IIoT, FL in IIoT may have a positive effect.

With the rapid popularization of IIoT technology, the geographical distribution of IIoT equipments is becoming more and more extensive and the equipment forms are quite different from each other \cite{a4}. This leads to a series of problems such as uneven data quality of equipment and high communication cost. In order to increase the model training rate of IIoT equipments and reduce the communication cost of model aggregation, driven by the current artificial intelligence (AI) technologies, we propose a deep reinforcement learning (DRL) assisted FL framework \cite{j2,9}. DRL algorithm is suitable for solving decision-making problems in high dimensions \cite{z4,10}, so it can use the decision-making effect of DRL to select some high-quality local IIoT equipment models for aggregation. Considering that a large amount of data generated by IIoT equipments will lead to excessive training data, increase the burden on wireless network channels and may cause privacy leaks, we employ a distributed method to train DRL agents, i.e., apply the DRL assisted FL method on each IIoT equipment to realize the effective application of FL in IIoT. The main work of this paper is as follows:
\begin{enumerate}
\item Aiming at the problem of difficult management of IIoT devices due to the generation of large amounts of heterogeneous and private data, this paper adopts a FL method to train and manage these data.

\item In order to improve the performance and efficiency of FL, this paper proposes a DRL-assisted FL framework. DRL algorithm based on deep deterministic strategy gradient (DDPG) is mainly used to select IIoT device nodes with high data quality, so as to increase the model aggregation rate and reduce communication costs.

\item We analyze the characteristics of IIoT device data, and then use MNIST, Fashion MNIST and CIFAR10 (IID and non-IID) data sets for experimental evaluation. Finally, we verify the effectiveness of FL technology based on DRL to process IIoT data.
\end{enumerate}

The remaining part of this paper is organized as follows: Section \ref{p2} reviews the work related to the application of AI technology in IIoT. Section \ref{p3} introduces in detail the related issues of the application of FL in IIoT. Section \ref{p4} introduces the realization of FL algorithm assisted by DRL and the node selection process of applying DRL algorithm. Section \ref{p5} describes the experimental setup of FL algorithm assisted by DRL and shows the experimental results. The last section concludes the paper.

\section{Related Work}\label{p2}

\subsection{Related Work of Industrial Internet of Things Based on Deep Reinforcement Learning}

IIoT has been widely concerned by the industry. Especially in the field of IIoT, there have been many solutions to solve the practical problems of IIoT by using AI technology. Due to the explosive growth of user equipments and data streams, IIoT has seen a shortage of spectrum resources. Shi et al. \cite{r1} proposed a spectrum resource management scheme for IIoT networks. Specifically, in order to consider the differentiated communication needs of different users, the authors proposed a modified deep Q-learning network (MDQN) and designed a new reward function to drive the learning process. Afterwards, the authors established a simple medium access control model, which used the base station as the sole agent to manage spectrum resources. In the end, the solution promoted the sharing of spectrum between different types of users. Chen et al. \cite{r2} studied the joint power control and dynamic resource management of multi-access edge computing (MEC) in IIoT. The authors transformed this problem into Markov decision process (MDP) and used the dynamic resource management algorithm based on DRL to solve this process. The algorithm fully considered the dynamic and continuity of task generation, and finally used the DDPG to optimize the long-term average delay of dynamic resource management. The experimental results showed that the method was effective. Based on blockchain technology, Liu et al. \cite{r3} proposed an emerging data collection and sharing scheme to deal with the two main problems faced by IIoT. The first was to achieve efficient data collection within the limited energy and sensing range of mobile terminals. The second was to ensure the security of data sharing between mobile terminals. The authors combined blockchain and DRL algorithms, used DRL to maximize data collection, and then used blockchain technology to ensure data security. This scheme embodied the superior performance of the two technologies.

\subsection{Related Work of Industrial Internet of Things Based on Federal Learning}

The latest development of FL in IIoT was an efficient deep anomaly detection framework based on equipment joint learning proposed by Liu et al. \cite{r4}, which was used to detect time series data in IIoT. The authors first used a FL framework to enable edge equipments to co-train an anomaly model. After that, they used the attention mechanism-based convolutional neural network (CNN) long short-term memory (LSTM) model to capture important fine-grained features. In the end, the authors used a gradient compression mechanism based on Top-k selection to improve communication efficiency. A large number of experiments on real data sets showed that the algorithm can effectively reduce communication overhead. FL was originally proposed by Brendan et al. \cite{r5}. They first proposed a decentralized method for private data training--FL, and then conducted extensive experimental evaluations using five different model structures and four data sets to verify the robustness of the FL method. This research opened a precedent for the exploration of FL technology. In addition, there were some typical representatives who applied FL to solve different actual network problems \cite{r6,r7,r8,zz2,zz3}, which had shown good results in solving the computing, caching and communication, and IIoT problems of the intelligent mobile edge.

We summarize the application classification of FL and DRL in TABLE \ref{tab_1}. Similar to the above works, we have jointly paid attention to the data privacy issues in the IIoT scene, and proposed feasible solutions from the perspective of design framework and algorithm implementation. However, the biggest difference between our work and the above works is that we use DRL algorithm to select high-quality IIoT equipment nodes for FL, so as to improve the rate of model aggregation and reduce the communication cost. This is not reflected in the above work.

\begin{table*}
\centering
\caption{Application classification combining FL and DRL.}
\renewcommand\arraystretch{1.5}
\begin{tabular}{|p{20mm}|p{45mm}|p{45mm}|p{45mm}|}
\hline
- & Advantage & Inferiority & Scale  \\
\hline
DRL & Best performance. & Data congestion; Security cannot be guaranteed. & The scale is small and can be applied to an edge node or UE. \\
\hline
Distribute DRL & Efficient training and learning; Unavailability at the edge. &  Security cannot be guaranteed; Performance is unstable. &  Medium-scale, suitable for resource allocation, computing offloading, caching strategy and other issues. \\
\hline
Federated Learning & Data security guarantee; Flexible training and learning process; Robust to non-IID data. & - & Large scale, suitable for issues such as resource allocation, computing offloading, caching strategy and traffic engineering. \\
\hline
\end{tabular}
\label{tab_1}
\end{table*}

\section{Federated Learning Application Problem Description in Industrial Internet of Things}\label{p3}

The core goal of FL is to carry out efficient ML among multi-UEs or multiple computing nodes under the premise of ensuring the security and privacy of data communication \cite{11}. IIoT equipments need to upload the local model to a central server after using local data for model training, and the central server will optimize the global model. The general steps are:

\begin{enumerate}
\item IIoT equipments use local data for local calculation, while minimizing the predefined empirical risk function, and then update the calculated weight to the wireless network access point.

\item The wireless network access point collects the weight of IIoT equipment update and accesses the FL unit to generate the global model.

\item FL redistributes the output of model training to IIoT equipments, which use global models for a new round of local training.

\item Repeat the above steps 2 and 3 until the loss function converges or reaches the maximum number of iterations.
\end{enumerate}

Suppose that there are $N$ IIoT equipment nodes in an IIoT scenario, the local data set composed of local data of each IIoT-DN is $\{X_1,X_2,...,X_N\}$. The IIoT equipment node downloads the global model $\theta$ from the central server and trains them by local data set alignment. IIoT equipment node uploads new weights or gradients to a central server to update the global model. Therefore, the data sample size from $N$ IIoT equipment node is $\sum_{n=1}^{N}x_n=X$. Then the loss function of IIoT equipment node with data set $X_N$ is

\begin{equation}
\begin{aligned}
F_n(\theta) \triangleq \frac{1}{x_n}\sum_{j \in X_n}f_j(\theta),
\end{aligned}
\end{equation}
where $f_j(\theta)$ is the loss function of data sample $j$. Optimize the global loss function by minimizing the weighted average of the local loss function $F_n(\theta)$ of each IIoT equipment node training sample:

\begin{equation}
\begin{aligned}
F(\theta) \triangleq \frac{\sum\limits_{j \in \cup_n} X_n f_j(\theta)}{|\cup_n X_n|} = \frac{\sum\limits_{n=1}^N X_n F_n(\theta)}{X}.
\end{aligned}
\end{equation}

We summarize the loss functions of several common ML models that can be used for FL in TABLE \ref{tab_2} \cite{a1}. DRL agent training real data from IIoT equipments has more advantages than agent data provided by the data center. The data generated by IIoT equipments is usually highly secretive and large in scale. DRL agents also need to record them on a central server when they are trained based on local models. In order to better verify the effectiveness of FL assisted by DRL, we extract the data characteristics generated by IIoT equipments as follows: (1) high privacy, (2) unbalanced amount of data (a lot of user output is used, the user output is low), (3) the distribution scale is large, and (4) the user equipment is limited by the communication quality. Corresponding to the actual situation of the data generated by IIoT equipments, our experiments are carried out on MNIST, Fashion MNIST and CIFAR-10 (independent and identically distributed (IID) and non-independent and identically distributed (non-IID)). The reason for using the above data sets is that they can reflect the characteristics of data heterogeneity, scale difference and dispersion to a certain extent.

In the case of non-IID, the data between users can be divided equally or unequal. We assume that there is a fixed number of IIoT devices in each round of communication between the server and devices. When the communication process starts, the devices are randomly divided into several groups, and the server will send the current global model parameters to each device. In order to improve the communication quality, we use DRL algorithm based on DDPG to select some device nodes to participate in the training. After that, each selected device will calculate according to the global state and local data set, and then send the update to the server. The server uses this repeated method to update the global model parameters. In this way, the non-IID characteristics of IIoT device data can be considered. The experimental setup part will be introduced in detail later.

\begin{table}
\centering
\caption{The loss function in several machine learning models.}
\renewcommand\arraystretch{1.5}
\begin{tabular}{|p{25mm}|p{55mm}|}
\hline
Model & Loss Function \\
\hline
Linear Regression & $\frac{1}{2}||y_i-w^Tx_i||^2$ \\
\hline
Logistic Regression & $\log(1+exp(-y_iw^Tx_i))$ \\
\hline
Smooth SVM & $\frac{1}{2}max\{0,1-y_iw^Tx_i\}$ \\
\hline
K-means & $\frac{1}{2}min_{j \in \{1,2,...,K'\}}||x_i-w_j||^2$, where $K'$ is the number of clusters. \\
\hline
Neural Network & $\frac{1}{2}||y_i-\sum_{n=1}^{N}v_n\phi(w_n^Tx_i)||$, where $v_n$ is the weights connecting the neurons, $N$ is the number of neurons and $\phi()$ is the activation function. \\
\hline
\end{tabular}
\label{tab_2}
\end{table}

\section{Implementation of Federated Learning Algorithm Assisted by DRL}\label{p4}

In this part, we will give a FL framework assisted by DRL in section \ref{p4-1}, which is the core technology to realize the data training of IIoT equipment. Based on this framework, we will describe in detail the implementation steps of the FL algorithm assisted by DRL in section \ref{p4-2}. Finally, we will introduce the process of applying DRL algorithm to achieve IIoT equipment selection.

\subsection{Framework}\label{p4-1}

The purpose of integrating DRL into FL is to intelligently use the cooperation between IIoT equipments and nodes to exchange learning parameters, so as to better train the local model \cite{12}. Due to the limited cache and performance calculation of the edge IIoT equipments and nodes, and the direct transmission of a large amount of data may cause network channel congestion or even data leakage, and so the data is trained in the local equipment. Therefore, we make full use of the efficient model training performance of DRL and conduct data confidentiality training based on FL. The final purpose is to verify the training effectiveness of equipment data in IIoT. We deploy DRL agents in every edge IIoT equipment. Considering the inherent characteristics of wireless network, if the amount of data to be trained is large, it will increase the burden on the wireless network channel. Moreover, the data needs to be transformed for privacy reasons, so the correlation of the central server-side agent data may not be as good as the data on the edge equipments. In addition, deploying DRL agents on edge equipments alone may cause additional energy consumption.

\begin{figure*}[!h]
\centering
\includegraphics[width=0.9\textwidth]{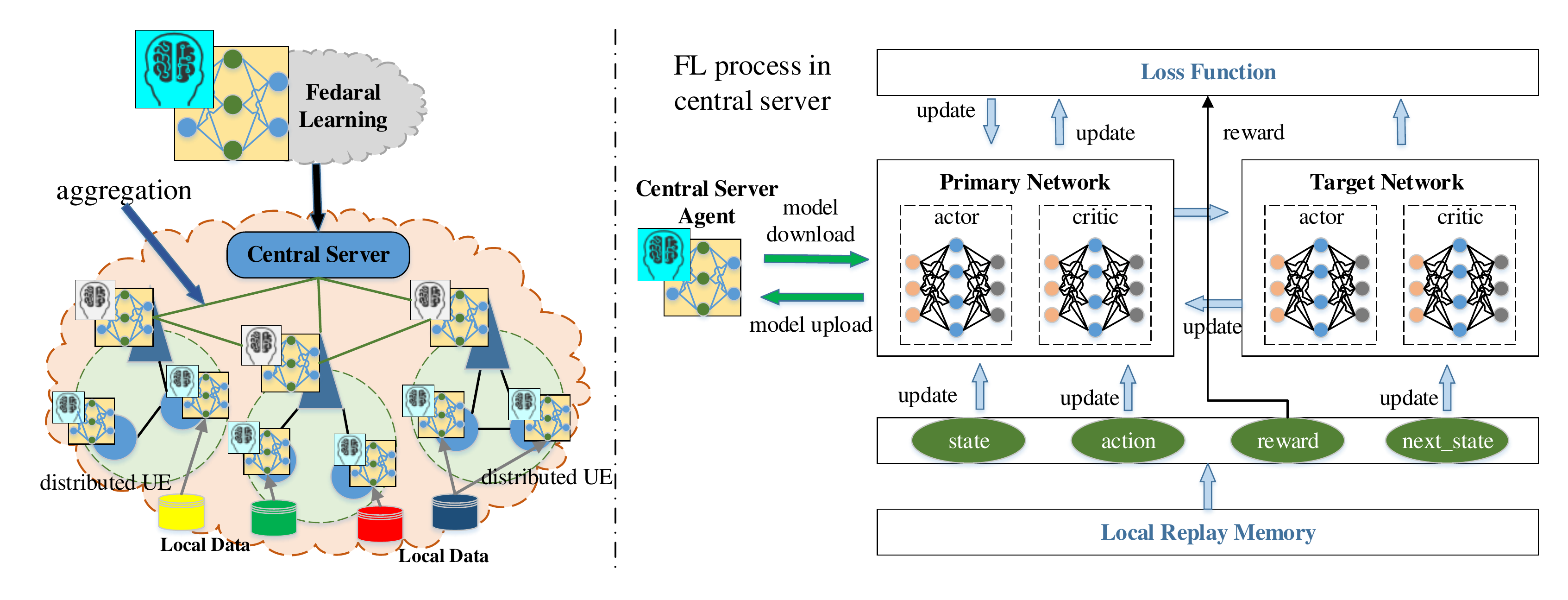}
\caption{Federated learning framework based on deep reinforcement learning.}
\label{fig_2}
\end{figure*}

To deal with the above problems, we propose a FL framework assisted by DRL, as shown in Fig. \ref{fig_2}. Considering a large number of data factors, we adopt the distributed deep learning (DL) method. As shown in TABLE \ref{tab_1}, distributed DL has good performance of fast training and edge efficiency. Therefore, the fast training of the data on edge equipments can avoid network congestion to certain extents. Considering that the correlation of agent data on the central server is not as good as that of the edge equipment, in the IIoT scenario, a large number of IIoT equipments are equipped with perceptron, which can obtain rich and personalized data to update the global model. Based on these data, the central server-side agent data can be effectively updated to match the data of the edge equipment side. The FL framework based on DRL mainly completes three tasks: node selection, local training and global aggregation. Node selection is achieved by using the DRL algorithm based on DDPG, in order to select IIoT nodes with high data quality to participate in the model training process. The purpose of local training is to derive local model parameters suitable for local nodes. The purpose of global aggregation is to generate a global model suitable for data training, which is achieved by uploading local model parameters to each local server.

\subsection{Implementation Steps}\label{p4-2}

The FL framework assisted by DRL has three main implementation stages:
\begin{enumerate}
\item Initialization stage: The central server evaluates the connection request of IIoT equipments. After that, the central server randomly selects a subset of user equipment from the connected IIoT equipments to participate in this round of training \cite{13,14}. After training, a global model $\theta_t$ is sent to each selected IIoT equipment.
\item Training stage: The selected IIoT equipment uses local data to train the global model, and the training process is $\theta_t \to \theta_t^n$, get the global model $\theta_t^{n+1}$ after each iteration. For the $n$-th IIoT equipment, the optimization objective of loss function is
\begin{equation}
\begin{aligned}
F(\theta) = \frac{\sum\limits_{n=1}^N X_n F_n(\theta)}{X},
\end{aligned}
\end{equation}
where $F(\theta)$ is the $n$-th local loss function.
\item Aggregation stage: All selected IIoT equipments upload the local training model to the central server, then the central server updates and trains the new global model $\theta_t^{n+1}$ for the next iteration. After that, the central server issues the new global model to the newly selected IIoT equipment collection. Repeat the above process until the loss function converges or reaches the maximum number of iterations.
\end{enumerate}

In the FL algorithm assisted by DRL that we designed, we select IIoT equipments with a proportion of $C$ in each round and then calculate the gradient loss of these IIoT equipment data. For IIoT equipment $n$, calculate the gradient $g_n=\nabla F_n(\theta_t)$ under the current model $\theta_t$ and then aggregate these gradients by the central server. Use the following formula to update:

\begin{equation}
\begin{aligned}
\theta_t - \alpha \sum\limits_{n=1}^N \frac{N_n}{N}g_n \to \theta_{t+1}.
\end{aligned}
\end{equation}

The operation of selecting specific IIoT devices is called a scheduling strategy. Since the reading and execution of data in the central server and the reading and execution of data in various IIoT devices are not in the same order of magnitude, uploading the training and updated models of all IIoT devices to the global controller will cause a lot of computing time and communication overhead. Therefore, it is an effective way to select a specific part of equipment upload parameter model through the scheduling strategy. Scheduling strategies play a crucial role in allocating wireless channels with limited resources to appropriate IIoT devices. Use $S=\frac{C}{N}$ to represent the ratio of the number of IIoT devices to the number of sub-channels. There are mainly three commonly used scheduling strategies:

\begin{enumerate}
\item Random scheduling strategy: In each round of communication, the central server randomly selects $N$ related IIoT devices for parameter update. At the same time, a dedicated sub-channel is built between the central server and the selected user equipment to transmit training parameters.

\item Cyclic scheduling strategy: The central server divides all IIoT devices into $G$ groups. Each time a group is selected to build a channel for communication, and the model parameters are updated cyclically during each communication.

\item Proportional fairness strategy: In each round of communication, select $N$ from $C$ related IIoT devices according to the following calculation method:
\end{enumerate}

\begin{equation}
\begin{aligned}
i^* = \mathop{\arg\max}\limits_{i \subset \{1,2,...,C\}} \{ \frac{\widetilde{\rho}_{i_1,t}}{\overline{\rho}_{i_1,t}},...,\frac{\widetilde{\rho}_{i_N,t}}{\overline{\rho}_{i_N,t}} \},
\end{aligned}
\end{equation}
where $i=\{i_1,i_2,...,i_N\}$ is a vector of length $N$. $i^*=\{i_i^*,i_2^*,...,i_N^*\}$ represents the index number of IIoT device. $\widetilde{\rho}_{i_i,t}$ and $\overline{\rho}_{i_i,t}$ represent the instantaneous and time average signal-to-noise ratios (SNRs) of IIoT device $i_n$ during the $t$-th communication, respectively.

The SNR received at the central server is expressed as follows:

\begin{equation}
\begin{aligned}
\gamma_{k,t}=\frac{P_{ut}h_k||c_k||^{-\beta}}{\sum\limits_{c \in \widetilde{\Psi}_{u}^{k}}P_{ut}h_c||c||^{-\beta}+\sigma^2},
\end{aligned}
\end{equation}
where $\beta$ represents the link loss index, $h_k$ represents the small-scale fading, $\sigma^2$ is the variance of Gaussian additive noise, and $\widetilde{\Psi}_{u}^{k}$ represents the position of IIoT device $k$.

An important purpose of model training is parameter update. At the number of communication rounds $t$, when the central server establishes a transmission channel with a certain IIoT equipment and successfully decodes the sent data, the parameter $\delta v_k^t$ can be updated from IIoT equipment to the central server. We use the probability of successful parameter update to indicate the transmission performance of the wireless channel, which is expressed as follows:

\begin{equation}
\begin{aligned}
U_k^b= P(\gamma_{k,t} > \theta, S_{k,t}^b=1),
\end{aligned}
\end{equation}
where $\gamma_{k,t}$ represents the instantaneous and time average signal-to-noise ratio at the central server. $S_{k,t}^b \in \{0,1\}$ represents the selection index, where $b$ represents different scheduling strategies. When $S_{k,t}^b=1$, it means that the communication time has started between the central server and IIoT equipment $k$, otherwise there is no communication.

\subsection{IIoT Equipment Node Selection Based on DRL}\label{p4-3}

Due to the different uses and geographical distribution of IIoT equipments, the data they generate will have the characteristics of heterogeneity, and the quality of the data will also vary. The problems of low data training efficiency, long model aggregation time and high wireless communication cost caused by data characteristics are major challenges in the application of FL in the field of IIoT. Therefore, in order to improve the quality of model aggregation and reduce communication costs, inspired by reference \cite{9}, we use DRL algorithm to select IIoT equipment nodes with accurate learning models. The main idea is to model the local training cost of IIoT equipments and describe the problem as a MDP. We use DDPG to find the optimal solution of MDP. DRL agent continuously interacts with the environment to accumulate maximum reward and the essential purpose is to minimize the cost of FL.

We first give the cost index for selecting IIoT equipment nodes. The total cost of node selection in FL algorithm is composed of training time cost and training quality cost. The training time cost of local IIoT equipment is
\begin{equation}
\begin{aligned}
C_{time}^t=\frac{\sum\limits_{i=1}^{num_{IIoT}}(c_l^t(i)+c_c^t(i))}{|num_{IIoT}|},
\end{aligned}
\end{equation}
where $C_{time}^t$ represents the total time cost of local training for all IIoT equipments, $c_l^t(i)$ represents the local training time cost of IIoT equipment $i$ in time slot $t$, and $c_c^t(i)$ represents the communication cost of IIoT equipment $i$ in time slot $t$.

In addition, we use the indicator of learning quality to characterize the loss of training accuracy:
\begin{equation}
\begin{aligned}
C_{qu}^t &=\sum_{i \in num_{IIoT}}\sigma_i^t(w^t,d_i) \\
&=\sum_{i \in num_{IIoT}}\sum_j Loss(y_j-\hat{w}^t(x_j)),
\end{aligned}
\end{equation}
where $\sigma_i=\sum_j Loss(y_j-\hat{y}_j)$ is used to quantify the quality of the network model, $w^t$ represents the training model aggregated in time slot $t$, and $d_i$ represents the training data of IIoT equipment $i$. Therefore, the total cost of DRL assisted FL algorithm in time slot $t$ is
\begin{equation}
\begin{aligned}
C^t=C_{time}^t+C_{qu}^t.
\end{aligned}
\end{equation}

The node selection of IIoT equipment is a combinatorial optimization problem. We model it as a MDP, denoted by $M=\{S,A,P_A,C_A\}$. Among them, $S$ represents the state space of IIoT equipment node, $A$ represents the action space, $P_A$ represents the state change probability of the action taken, and $C_A$ represents the cost of the new state produced by the action. Therefore, the selection problem of IIoT equipment nodes is expressed as:

\begin{equation}
\begin{aligned}
\mathop{\min}\limits_{\delta^t} C^t(\delta^t),
\end{aligned}
\end{equation}
where $\delta^t$ represents the selection status of the IIoT equipment node. When the IIoT equipment node $i$ is selected, $\delta_i^t=1$, otherwise $\delta_i^t=0$.

The DRL agent trains the local model by interacting with the environment, and then we use the DDPG to select the optimal solution for IIoT equipment node. In the DRL algorithm, we use the reward function to evaluate the effect of taking action $a_t$. The evaluation method is as follows:

\begin{equation}
\begin{aligned}
r(s_t,a_t) &=-\frac{\sum\limits_{i=1}^{n}C_i^t \cdot a_i^t}{\sum\limits_{i=1}^{n}a_i} \\
&=-\frac{(\sum\limits_{i=1}^{n}a_i(\frac{d_i \cdot T_m}{\mu_i(t)}+\frac{w_i}{\tau_i})+\sum\limits_{i=1}^{n}a_i\sigma_i^t(w_i^t,d_i))}{\sum\limits_{i=1}^{n}a_i},
\end{aligned}
\end{equation}
where $T_m$ represents the number of CPU cycles required to train the model $m$ on the data $d_i$, $\mu_i$ represents the computing resources available to IIoT equipment $i$, and $\tau$ represents the transmission rate of the wireless network channel.

The DDPG uses a value function to determine the strategy, which mainly includes an actor deep neural network (DNN) $\pi(s_t|\theta_{pi})$ and a critic DNN $Q(s_t,a_t|\theta_Q)$. The DDPG uses historical experience stored in local replay memory to perform information conversion. The converted information includes the current state $s_t$, the action taken $a_t$, the next state $s_{t+1}$ and the reward $r(s_t,a_t)$ obtained by taking the action $a_t$. The target network generates target values to train the critic DNN model.

The specific pseudo code is given in Algorithm 1, where the client label is indexed by $n$, $B$ represents the local mini batch size, $E$ represents the number of local epoch, $\alpha$ represents the learning rate.

\begin{algorithm}[h]
  \caption{Federated learning algorithm assisted by deep reinforcement learning}
  \begin{algorithmic}[1]
    \Require
    {$\theta_\pi,\theta_Q$};
    \State
    {$Set \,\, \theta_\pi^{tar} = \theta_\pi \,\, and \,\, \theta_Q^{tar} = \theta_Q$};
    \State
    {$Initialize \,\, actor \,\, DNN \,\, and \,\, critic \,\, DNN \,\, parameters$};
    \State {$K \,\, clients, \,\, B \,\, batch \,\, size, \,\, \alpha \,\, learning \,\, rate$};
    \State $\textbf{Server\,\,executs:}$
    \State $initialize \,\, \theta_0$;
    \For{each round}
      \State $max(C \cdot K, 1) \to \textit{num};$
      \State $(random\_clients(\textit{num})) \to S_t;$
      \State{$select\,\,client \,\, n \,\, \in \,\, S_t$};
        \State $ClientUpdate(n,\theta_t) \to \theta_{t+1}^n;$
      \State $\sum\limits_{n=1}^{N} \frac{N_n}{N}g_n \to \theta_{t+1}^{n};$
    \EndFor
    \State $\textbf{ClientUpdate($n,\theta$):}$
      \State $(split \,\, P_n \,\, into \,\, batches \,\, of \,\, size \,\, B) \to \beta;$
      \For{$each \,\, local \,\, epoch$}
        \For{$batch \,\, b \in \beta$}
            \State $\theta - \alpha \nabla \zeta (\theta;b);$
            \EndFor
      \EndFor
    \State $\textbf{DDPG:}$
      \For{$each \,\, episode$}
        \For{$time \,\, slot \,\, t$}
          \State $take \,\, action \,\, a_t;$
          \State $caculate \,\, r(s_t,a_t) \,\, and \,\, update \,\, s_t \,\, to \,\, s_{t+1};$
          \State $Sample \,\, a \,\, mini \,\, batch \,\, of \,\, experiences$
          \State $\qquad\qquad from \,\, local \,\, replay \,\, memory;$
          \State $update \,\, \pi(s|\theta_{\pi}) \,\, and \,\, Q(s,a|\theta_Q);$
          \State $update \,\, the \,\, target \,\, network \,\, parameters$
          \State $\qquad\qquad to \,\, local \,\, replay \,\, memory;$
          \State $store \,\, the \,\, new \,\, experiences;$
        \EndFor
      \EndFor
    \State $return \,\, \theta \,\, to \,\, server;$
  \end{algorithmic}
\end{algorithm}

\section{Experimental Setup and Result Analysis}\label{p5}

In this part, we evaluate the performance of FL algorithm assisted by DRL based on MNIST, Fashion MNIST and CIFAR-10 (IID and non-IID). Since the above data sets reflect the characteristics of IIoT equipment data, we mainly test the accuracy of the algorithm in training the above data sets. This shows that the FL algorithm assisted by DRL has advantages in training IIoT equipment data.

\subsection{Experimental Setup}

Considering the host performance, we mainly use several small agent data sets to test the algorithm performance. MNIST data set and Fashion MNIST data set are mainly used for number recognition tasks. The former uses a simple multi-layer perception (MLP) with two hidden layers, in which each layer has 200 units and is activated by the ReLU function \cite{15}. The latter is trained by CNN, which has a fully connected layer (512 units and ReLU function), a softmax output layer and two $5 \times 5$ convolution layers. To fully study FL, we use two data partitioning methods for IIoT equipments. The first is the IID of data shuffling and then divided into 100 equipment receiving 600 data samples respectively. The second is the non-IID which classifies the data according to the digital tags, which is divided into 200 pieces of 300 pieces and each client is allocated 2 pieces. This setting method can explore the breaking degree of the algorithm on highly non-IID data. CIFAR-10 data consists of 10 classes of $32 \times 32$ images with three GRB channels. We used 50,000 training cases and 10,000 test cases and distributed them equally to 100 equipment.

\subsection{Performance Evaluation}

We first explore the impact of the number of IIoT equipment, i.e., the number of clients, on the MNIST data set and fashion MNIST data set models. The client ratio $C$ controls the number of multi client parallelism. Because we use 100 IIoT device as clients, when $C = 0$, it represents one client, followed by 10, 20, 50 and 100 clients in turn. TABLE \ref{tab_3} shows the number of communication rounds that different clients account for to reach the required precision for training the two data sets.

\begin{table}
\centering
\caption{The influence of the proportion $C$ of different clients on the MNIST data set with $E=1$ and the Fashion MNIST data set with $E=5$.}
\begin{threeparttable}
\renewcommand\arraystretch{1.5}
\begin{tabular}{|p{12mm}|p{5mm}|p{10mm}|p{10mm}|p{10mm}|p{10mm}|}
\hline
\multirow{2}{*}{} & \multirow{2}{*}{$C$} & \multicolumn{2}{c|}{IID} & \multicolumn{2}{c|}{non-IID} \\
\cline{3-6}
& \multirow{2}{*}{} & $B=\infty$ & $B=10$ & $B=\infty$ & $B=10$ \\
\hline
\multirow{5}{*}{\makecell[l]{MNIST \\ \\ $E=1$}} & 0 & 1455 & 316 & 4278 & 3275 \\
\cline{2-6}
& 0.1 & 1474 & 87 & 1796 & 664 \\
\cline{2-6}
& 0.2 & 1658 & 71 & 1528 & 619 \\
\cline{2-6}
& 0.5 & - & 75 & - & 443 \\
\cline{2-6}
& 1.0 & - & 70 & - & 380 \\
\hline
\multirow{5}{*}{\makecell[l]{Fashion \\ MNIST \\ \\ $E=5$}} & 0 & 387 & 50 & 1181 & 956 \\
\cline{2-6}
& 0.1 & 339 & 18 & 1100 & 206 \\
\cline{2-6}
& 0.2 & 337 & 18 & 978 & 200 \\
\cline{2-6}
& 0.5 & 164 & 18 & 1067 & 261 \\
\cline{2-6}
& 1.0 & 246 & 16 & - & 97 \\
\hline
\end{tabular}
 \begin{tablenotes}
        \footnotesize
        \item Unit: Number of communication rounds.
      \end{tablenotes}
  \end{threeparttable}
\label{tab_3}
\end{table}

\begin{figure}[!h]
\centering
\includegraphics[width=1\columnwidth]{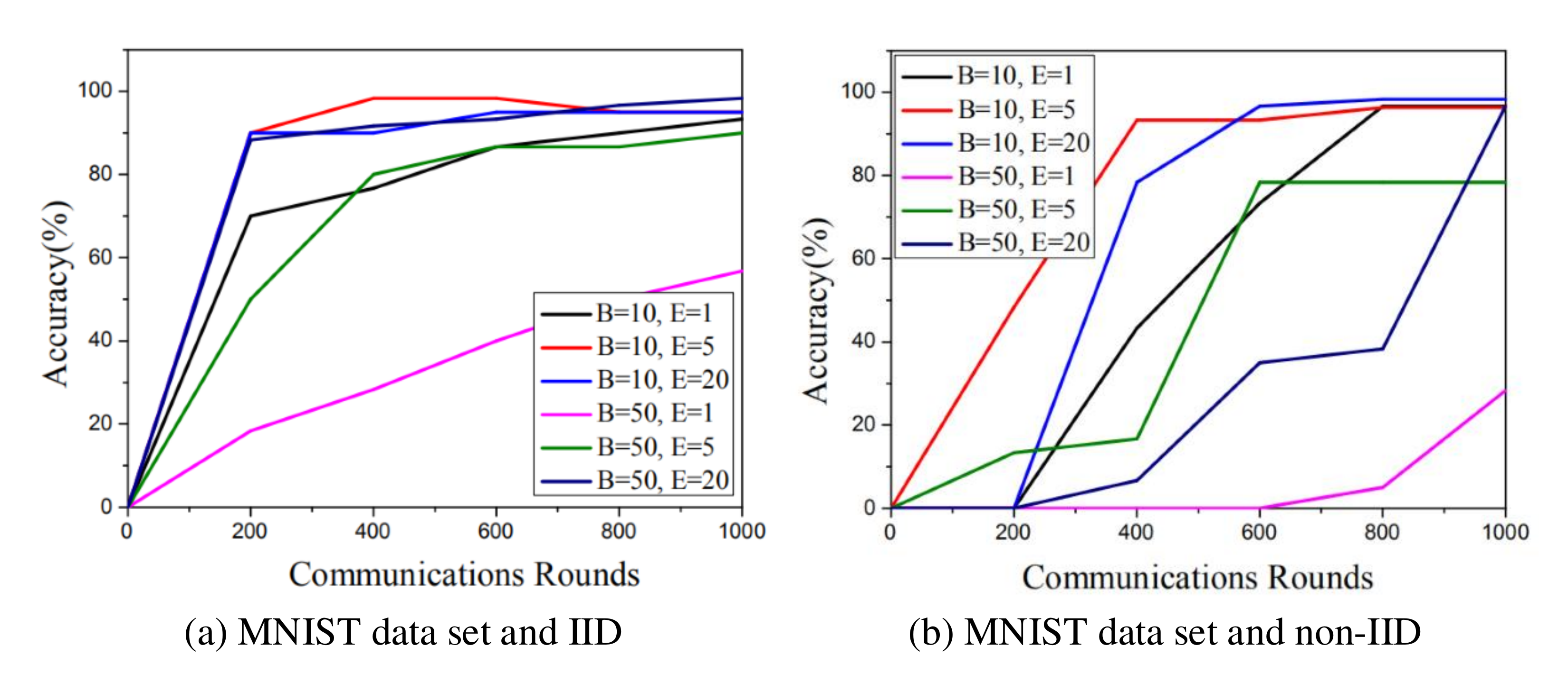}
\caption{Experimental results of MNIST data set.}
\label{fig_3}
\end{figure}

\begin{figure}[!h]
\centering
\includegraphics[width=1\columnwidth]{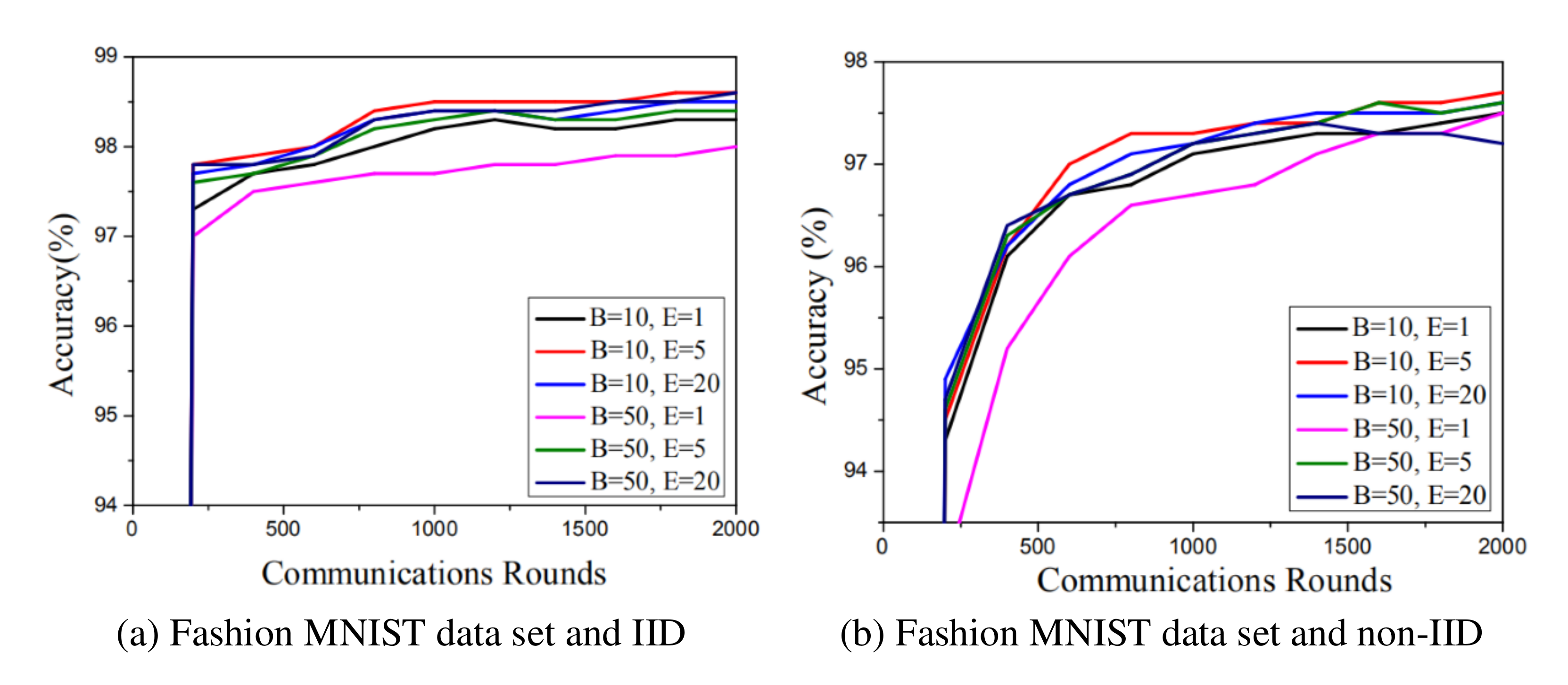}
\caption{Experimental results of Fashion MNIST data set.}
\label{fig_4}
\end{figure}

When $B = \infty$ is used for data training, all 600 data samples are treated as a single batch processing in each round. At this time, increasing the proportion of clients can only show a small advantage. When $B = 10$ is used for batch processing (especially when $C \ge 0.1$), the advantage of communication round number change is obvious. In order to balance the computational efficiency and the convergence rate, we fix $C = 0.1$. Figs. \ref{fig_3} and \ref{fig_4} show the gradual improvement of the accuracy of MNIST data set and Fashion MNIST data set after different training rounds. Each data set is trained based on IID data and non-IID data, respectively. It can be seen that after about 1,000 rounds of communication, the accuracy rate of MNIST data set can reach 99\% and then it tends to be stable. After about 2,000 rounds of communication, the training accuracy of the Fashion MNIST data set reached 97\% and stabilized.

\begin{figure}[!h]
\centering
\includegraphics[width=1\columnwidth]{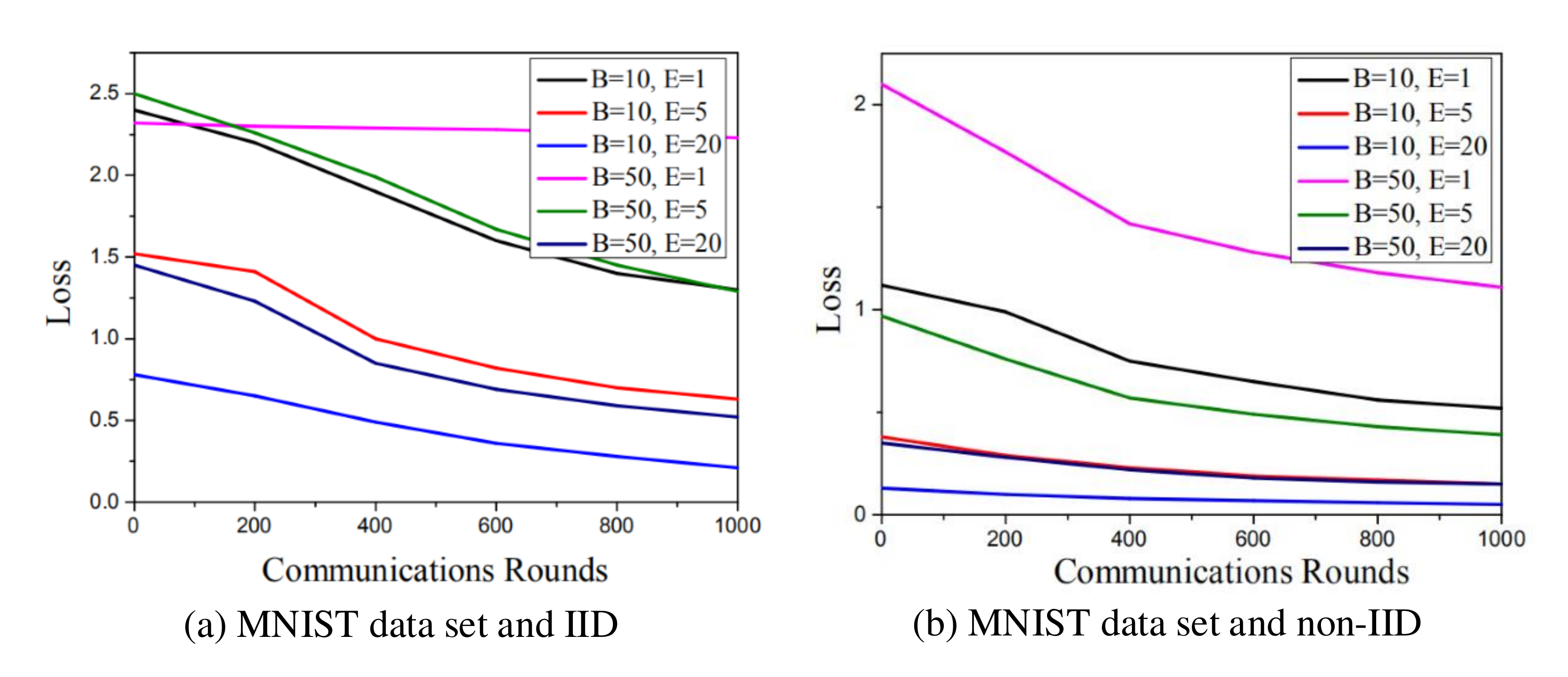}
\caption{The convergence of the loss value.}
\label{fig_5}
\end{figure}

Fig. \ref{fig_5} shows the convergence of the loss value during the training of MNIST data set. The loss value of MNIST data set based on IID data and non-IID data decreases and converges with the increase of communication rounds. After about 1,000 rounds of communication, the loss value is stable at a low level, which shows the rationality of the loss function and the effectiveness of the training method. The experimental results of training loss value on Fashion MNIST data set show that the loss value of fashion MNIST data set converges with the increase of communication rounds, which is similar to MNIST data set.

\begin{figure}[!h]
\centering
\includegraphics[width=0.95\columnwidth]{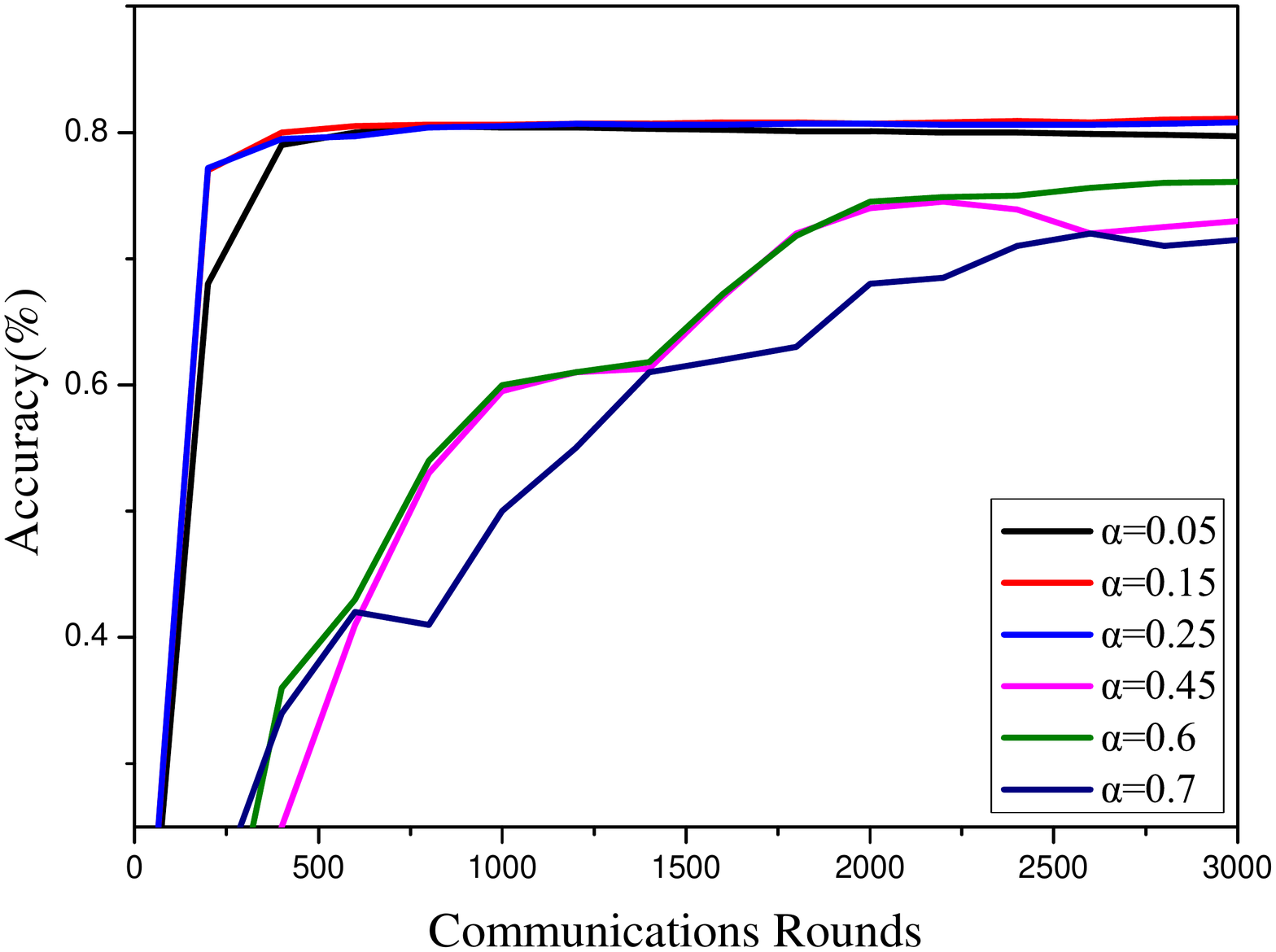}
\caption{Experimental results of CIFAR-10 data set based on IID data. Fixed $B = 50$, $E = 5$.}
\label{fig_6}
\end{figure}

Since the CIFAR-10 data set has no natural user partition, we consider the balancing and IID settings. The architecture of the training model is derived from Tensorflow, which mainly includes two convolution layers, two full connection layers and one linear conversion layer. In the process of training, we use stochastic gradient descent (SGD) method for small batch training with size of 100. Fig. \ref{fig_6} shows the relationship between the test accuracy of CIFAR-10 data set and the number of communication rounds.

Through the above experiments, we can see the advantages of the FL algorithm assisted by DRL in training the data of IIoT equipment. Considering the outstanding characteristics of heterogeneity and privacy of data generated by IIoT equipments, we use MNIST, Fashion MNIST and CIFAR-10 data sets to represent them. Based on MLP and CNN to train the above data sets, the training accuracy is rising in the process of continuous communication between the client and the central server. Among them, CIFAR-10 data set can achieve 85\% test accuracy, while MNIST and Fashion MNIST data sets can achieve more than 98\% training accuracy. Therefore, it can be shown that the FL algorithm assisted by DRL has excellent performance and has the ability to effectively manage and train the equipment data of IIoT.

In addition, we compare the algorithm proposed in this paper with a baseline algorithm called FedSGD \cite{two2}. In each round of communication, we select IIoT devices with a proportion of $C$, and then calculate the loss gradient of these devices. We fix the value of $C$ to 0.1, and then discuss the number of communication rounds required for the algorithm to achieve the target accuracy when $B$ or $E$ changes. The results are shown in TABLE \ref{tab_4}.

\begin{table}
\centering
\caption{Number of communication rounds required to achieve 98\% accuracy.}
\renewcommand\arraystretch{1.5}
\begin{tabular}{|p{20mm}|p{7mm}|p{7mm}|p{10mm}|p{10mm}|}
\hline
\multicolumn{5}{|c|}{MNIST $C=0.1$} \\
\hline
Algorithm & $E$ & $B$ & IID & non-IID \\
\hline
FedSGD & 1 & $\infty$ & 625 & 484 \\
\hline
Our algorithm & 20 & $\infty$ & 235 & 672 \\
\hline
Our algorithm & 5 & $\infty$ & 179 & 1000 \\
\hline
Our algorithm & 1 & 50 & 65 & 598 \\
\hline
Our algorithm & 1 & 10 & 34 & 350 \\
\hline
Our algorithm & 20 & 50 & 32 & 423 \\
\hline
Our algorithm & 5 & 10 & 20 & 229 \\
\hline
Our algorithm & 20 & 10 & 17 & 173 \\
\hline
\end{tabular}
\label{tab_4}
\end{table}

It can be seen from the table that the change of $E$ value and $B$ value has obvious influence on the number of communication rounds. We set the target accuracy to 98\%. For IID data, increasing the value of $E$ and reducing the value of $B$ can significantly reduce the number of communication rounds to reach this accuracy. But for non-IID data, although the number of communication rounds will be reduced, the overall effect is not obvious. In general, the larger $E$ is and the smaller $B$ is, the optimization effect is more obvious. So overall, the algorithm proposed in this paper is effective in improving the data training rate. Each selected IIoT device calculates the average gradient of local data based on the current model, and then the central server aggregates these gradients and updates them using equation (4). Each IIoT device uses its local data to perform one-step gradient descent based on the current model, and then the server performs a weighted average of the generated model. The final algorithm achieved good results.

\section{Conclusion}\label{p6}

IIoT is the key technology to realize industry 4.0 and it is also the objective embodiment of the development degree of industry 4.0. What cannot be ignored is that the current IIoT is facing the management and training problems brought by the explosive growth of user data. The emergence of FL provides a new solution paradigm for heterogeneous data and private data training, and it can support the development of IIoT in a way that reduces model deviation. This paper mainly aims at the problem of how to manage and train a large amount of data produced by IIoT, and proposes a FL algorithm assisted by DRL in wireless network environments. DRL based on DDPG is mainly used for the selection of IIoT equipment nodes. We fully analyze the heterogeneity and privacy of the data generated by IIoT equipments. In the experimental phase, we use MNIST, Fashion MNIST and CIFAR-10 data sets to represent IIoT equipment data. The final results show that the FL algorithm assisted by DRL can effectively train the above data sets and achieve a high accuracy rate, which shows the effectiveness of the FL algorithm assisted by DRL in the management of IIoT equipment data.

\ifCLASSOPTIONcaptionsoff
  \newpage
\fi

\section*{Biographies}

\begin{IEEEbiography}[{\includegraphics[width=1in,height=1.25in,clip,keepaspectratio]{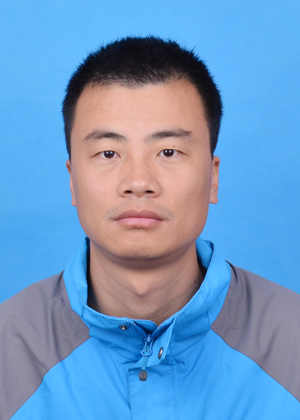}}]{Peiying Zhang}
is currently an Associate Professor with the College of Computer Science and Technology, China University of Petroleum (East China). He received his Ph.D. in the School of Information and Communication Engineering at University of Beijing University of Posts and Telecommunications in 2019. He has published multiple IEEE/ACM Trans./Journal/Magazine papers since 2016, such as IEEE TVT, IEEE TNSE, IEEE TNSM, IEEE TETC, IEEE Network, IEEE Access, IEEE IoT-J, ACM TALLIP, COMPUT COMMUN, IEEE COMMUN MAG, and etc. He served as the Technical Program Committee of ISCIT 2016, ISCIT 2017, ISCIT 2018, ISCIT 2019, Globecom 2019, COMNETSAT 2020, SoftIoT 2021, IWCMC-Satellite 2019, and IWCMC-Satellite 2020. His research interests include semantic computing, future internet architecture, network virtualization, and artificial intelligence for networking.
\end{IEEEbiography}

\begin{IEEEbiography}[{\includegraphics[width=1in,height=1.25in,clip,keepaspectratio]{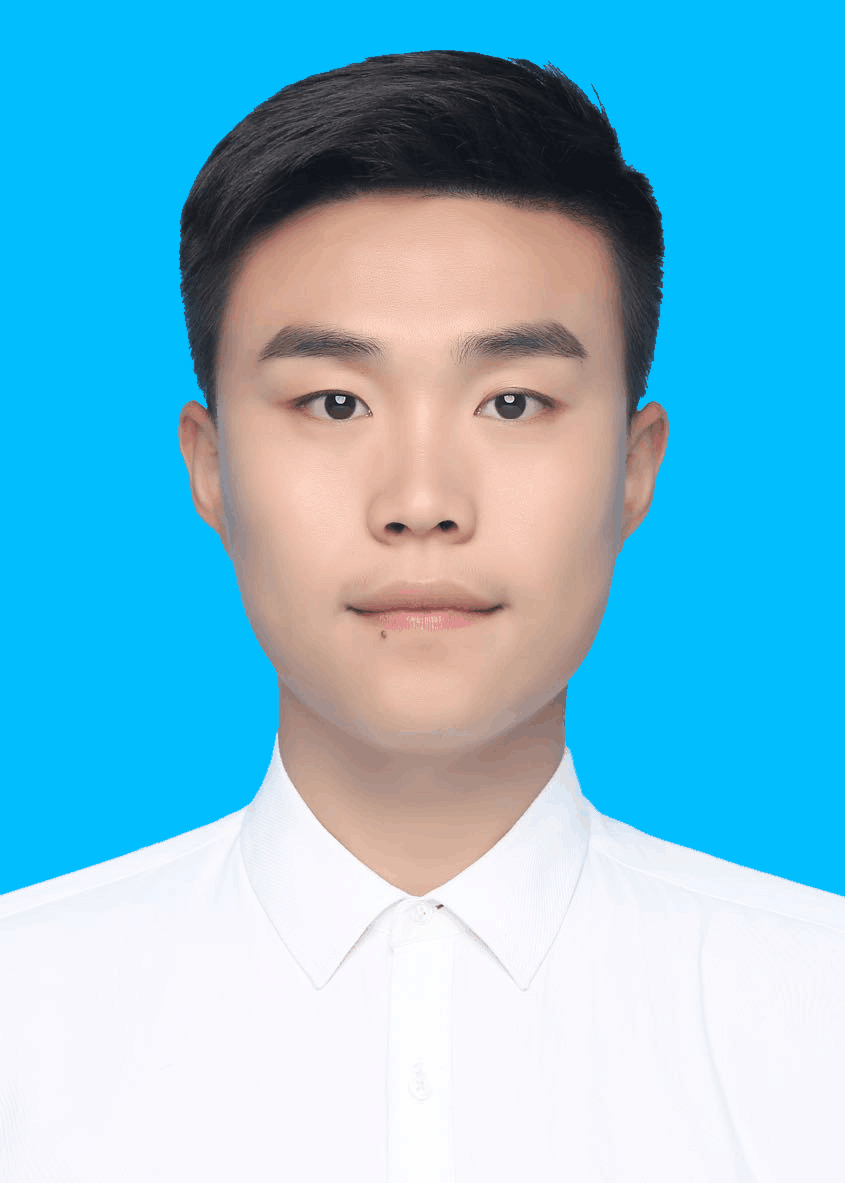}}]{Chao Wang}
is currently studying for a master's degree in Computer Science and Technology at the School of Computer Science and Technology, China University of Petroleum (East China), majoring in computer technology. He received his bachelor's degree in 2019. He has published several high-level papers as the first author or participating author, including IEEE TVT, IEEE TNSE, Software Practice \& Experience and Computing. He has also participated in important international conferences such as WoWMoM 2020 and ComComAp 2020. His research interests include virtual network embedded algorithms, network artificial intelligence, deep reinforcement learning, future network architecture, Internet of Things technology and wireless network communication.
\end{IEEEbiography}

\begin{IEEEbiography}[{\includegraphics[width=1in,height=1.25in,clip,keepaspectratio]{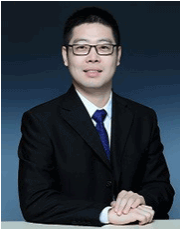}}]{Chunxiao Jiang}
is an associate professor in School of Information Science and Technology, Tsinghua University. He received the B.S. degree in information engineering from Beihang University, Beijing in 2008 and the Ph.D. degree in electronic engineering from Tsinghua University, Beijing in 2013, both with the highest honors. Dr. Jiang has served as an Editor of IEEE Internet of Things Journal, IEEE Network, IEEE Communications Letters, and a Guest Editor of IEEE Communications Magazine, IEEE Transactions on Network Science and Engineering and IEEE Transactions on Cognitive Communications and Networking. He has also served as a member of the technical program committee as well as the Symposium Chair for a number of international conferences. His research interests include application of game theory, optimization, and statistical theories to communication, networking, and resource allocation problems, in particular space networks and heterogeneous networks.
\end{IEEEbiography}

\begin{IEEEbiography}[{\includegraphics[width=1in,height=1.25in,clip,keepaspectratio]{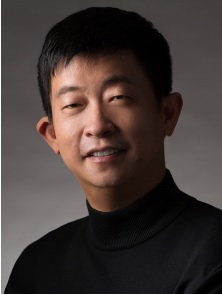}}]{Zhu Han}
received the B.S. degree in electronic engineering from Tsinghua University, in 1997, and the M.S. and Ph.D. degrees in electrical and computer engineering from the University of Maryland, College Park, in 1999 and 2003, respectively. From 2000 to 2002, he was an R\&D Engineer of JDSU, Germantown, Maryland. From 2003 to 2006, he was a Research Associate with the University of Maryland. From 2006 to 2008, he was an Assistant Professor with Boise State University, Idaho. He is currently a John and Rebecca Moores Professor with the Electrical and Computer Engineering Department as well as in the Computer Science Department, the University of Houston, Texas, and also with the Department of Computer Science and Engineering, Kyung Hee University, Seoul, South Korea His research interests include wireless resource allocation, wireless communications, game theory, big data analysis, security, and smart grid.
\end{IEEEbiography}


\begin{thebibliography}{1}

\bibitem{1}
M. Aazam, S. Zeadally and K. A. Harras, ``Deploying Fog Computing in Industrial Internet of Things and Industry 4.0," {\em IEEE Transactions on Industrial Informatics}, vol. 14, no. 10, pp. 4674-4682, Oct. 2018.

\bibitem{2}
T. Wang, H. Luo, W. Jia, A. Liu and M. Xie, ``MTES: An Intelligent Trust Evaluation Scheme in Sensor-Cloud-Enabled Industrial Internet of Things," {\em IEEE Transactions on Industrial Informatics}, vol. 16, no. 3, pp. 2054-2062, Mar. 2020.

\bibitem{two4}
L. Cui, Z. Chen, S. Yang, Z. Ming, Q. Li, Y. Zhou, S. Chen and Q. Lu, ``A Blockchain-based Containerized Edge Computing Platform for the Internet of Vehicles,'' {\em IEEE Internet of Things Journal}, pp. 1-15, 2020, doi: 10.1109/JIOT.2020.3027700.

\bibitem{3}
M. Aazam, K. A. Harras and S. Zeadally, ``Fog Computing for 5G Tactile Industrial Internet of Things: QoE-Aware Resource Allocation Model," {\em IEEE Transactions on Industrial Informatics}, vol. 15, no. 5, pp. 3085-3092, May 2019.

\bibitem{two3}
W. Zhang, W. Guo, X. Liu, Y. Liu, J. Zhou, B. Li, Q. Lu and S. Yang, ``LSTM-Based Analysis of Industrial IoT Equipment,'' {\em IEEE Access}, vol. 6, pp. 23551-23560, 2018.

\bibitem{jcx1}
C. Jiang, Y. Chen, K. J. R. Liu and Y. Ren, ``Renewal-Theoretical Dynamic Spectrum Access in Cognitive Radio Network with Unknown Primary Behavior,'' {\em IEEE Journal on Selected Areas in Communications}, vol. 31, no. 3, pp. 406-416, 2013.

\bibitem{jcx2}
C. Jiang, Y. Chen, Y. Gao and K. J. R. Liu, ``Joint Spectrum Sensing and Access Evolutionary Game in Cognitive Radio Networks,'' {\em IEEE Transactions on Wireless Communications}, vol. 12, no. 5, pp. 2470-2483, 2013.

\bibitem{jcx3}
X. Zhu, C. Jiang, L. Kuang, N. Ge and J. Lu, ``Non-orthogonal Multiple Access Based Integrated Terrestrial-Satellite Networks,'' {\em IEEE Journal on Selected Areas in Communications}, vol. 35, no. 10, pp. 2253-2267, Oct. 2017.

\bibitem{j1}
J. Wang, C. Jiang, K. Zhang, X. Hou, Y. Ren and Y. Qian, ``Distributed Q-Learning Aided Heterogeneous Network Association for Energy-Efficient IIoT," {\em IEEE Transactions on Industrial Informatics}, vol. 16, no. 4, pp. 2756-2764, Apr. 2020.

\bibitem{a3}
W. Zhang, Q. Lu, Q. Yu, Z. Li, Y. Liu, S. K. Lo, S. Chen, X. Xu and L. Zhu, ``Blockchain-based Federated Learning for Device Failure Detection in Industrial IoT," {\em IEEE Internet of Things Journal}, pp. 1-12, 2020, doi: 10.1109/JIOT.2020.3032544.

\bibitem{zz1}
Y. Dai, D. Xu, K. Zhang, S. Maharjan and Y. Zhang, ``Deep Reinforcement Learning and Permissioned Blockchain for Content Caching in Vehicular Edge Computing and Networks,'' {\em IEEE Transactions on Vehicular Technology}, vol. 69, no. 4, pp. 4312-4324, Apr. 2020.

\bibitem{5}
M. I. Aziz Zahed, I. Ahmad, D. Habibi and Q. V. Phung, ``Content Caching in Industrial IoT: Security and Energy Considerations," {\em IEEE Internet of Things Journal}, vol. 7, no. 1, pp. 491-504, Jan. 2020.

\bibitem{z3}
P. Zhang, C. Wang, C. Jiang and A. Benslimane, ``Security-Aware Virtual Network Embedding Algorithm based on Reinforcement Learning," {\em IEEE Transactions on Network Science and Engineering}, pp. 1-11, 2020, doi: 10.1109/TNSE.2020.2995863.

\bibitem{6}
F. Sattler, S. Wiedemann, K. R. Muller and W. Samek, ``Robust and Communication-Efficient Federated Learning From Non-i.i.d. Data," {\em IEEE Transactions on Neural Networks and Learning Systems}, vol. 31, no. 9, pp. 3400-3413, Sep. 2020.

\bibitem{7}
N. I. Mowla, N. H. Tran, I. Doh and K. Chae, ``AFRL: Adaptive federated reinforcement learning for intelligent jamming defense in FANET," {\em Journal of Communications and Networks}, vol. 22, no. 3, pp. 244-258, Jun. 2020.

\bibitem{8}
Y. Lu, X. Huang, Y. Dai, S. Maharjan and Y. Zhang, ``Differentially Private Asynchronous Federated Learning for Mobile Edge Computing in Urban Informatics," {\em IEEE Transactions on Industrial Informatics}, vol. 16, no. 3, pp. 2134-2143, Mar. 2020.

\bibitem{two1}
F. Liu, X. Wu, S. Ge, W. Fan and Y. Zou, ``Federated Learning for Vision-and-Language Grounding Problems,'' {\em Proceedings of the AAAI Conference on Artificial Intelligence}, vol. 34, no. 7, pp: 11572-11579, 2020.

\bibitem{a4}
S. K. Lo, Q. Lu, C. Wang and H. Y. Paik, ``A Systematic Literature Review on Federated Machine Learning: From A Software Engineering Perspective," vol. abs/2007.11354, July 2020, https://arxiv.org/abs/2007.11354v1.

\bibitem{j2}
C. Jiang and X. Zhu, ``Reinforcement Learning Based Capacity Management in Multi-Layer Satellite Networks," {\em IEEE Transactions on Wireless Communications}, vol. 19, no. 7, pp. 4685-4699, Jul. 2020.

\bibitem{9}
Y. Lu, X. Huang, K. Zhang, S. Maharjan and Y. Zhang, ``Blockchain Empowered Asynchronous Federated Learning for Secure Data Sharing in Internet of Vehicles," {\em IEEE Transactions on Vehicular Technology}, vol. 69, no. 4, pp. 4298-4311, Apr. 2020.

\bibitem{z4}
P. Zhang, H. Yao and Y. Liu, ``Virtual Network Embedding Based on Computing, Network, and Storage Resource Constraints," {\em IEEE Internet of Things Journal}, vol. 5, no. 5, pp. 3298-3304, Oct. 2018.

\bibitem{10}
J. Wang, C. Jiang, H. Zhang, Y. Ren, K. C. Chen and L. Hanzo, ``Thirty Years of Machine Learning: The Road to Pareto-Optimal Wireless Networks," {\em IEEE Communications Surveys \& Tutorials}, vol. 22, no. 3, pp. 1472-1514, Jul. 2020.

\bibitem{r1}
Z. Shi, X. Xie, H. Lu, H. Yang, M. Kadoch and M. Cheriet, ``Deep Reinforcement Learning based Spectrum Resource Management for Industrial Internet of Things," {\em IEEE Internet of Things Journal}, pp. 1-14, 2020, doi: 10.1109/JIOT.2020.3022861.

\bibitem{r2}
Y. Chen, Z. Liu, Y. Zhang, Y. Wu, X. Chen and L. Zhao, ``Deep Reinforcement Learning based Dynamic Resource Management for Mobile Edge Computing in Industrial Internet of Things," {\em IEEE Transactions on Industrial Informatics}, pp. 1-9, 2020, doi: 10.1109/TII.2020.3028963.

\bibitem{r3}
C. H. Liu, Q. Lin and S. Wen, ``Blockchain-Enabled Data Collection and Sharing for Industrial IoT With Deep Reinforcement Learning," {\em IEEE Transactions on Industrial Informatics}, vol. 15, no. 6, pp. 3516-3526, Jun. 2019.

\bibitem{r4}
Y. Liu, S. Garg, J. Nie, Y. Zhang and Z. Xiong, ``Deep Anomaly Detection for Time-series Data in Industrial IoT: A Communication-Efficient On-device Federated Learning Approach," {\em IEEE Internet of Things Journal}, pp. 1-11, 2020, doi: 10.1109/JIOT.2020.3011726.

\bibitem{r5}
H. B. McMahan, E. Moore, D. Ramage, S. Hampson and A. B. Aguera y, ``Communication-efficient learning of deep networks from decentralized data," {\em 20th International Conference on Artificial Intelligence and Statistics, Fort Lauderdale, FL}, Apr. 2017.

\bibitem{r6}
X. Wang, Y. Han, C. Wang, Q. Zhao, X. Chen and M. Chen, ``In-Edge AI: Intelligentizing Mobile Edge Computing, Caching and Communication by Federated Learning," {\em IEEE Network}, vol. 33, no. 5, pp. 156-165, Sep.-Oct. 2019.

\bibitem{r7}
Q. Wu, K. He and X. Chen, ``Personalized Federated Learning for Intelligent IoT Applications: A Cloud-Edge Based Framework," {\em IEEE Open Journal of the Computer Society}, vol. 1, pp. 35-44, May 2020.

\bibitem{r8}
J. Kang, Z. Xiong, D. Niyato, Y. Zou, Y. Zhang and M. Guizani, ``Reliable Federated Learning for Mobile Networks," {\em IEEE Wireless Communications}, vol. 27, no. 2, pp. 72-80, Apr. 2020.

\bibitem{zz2}
T. T. Anh, N. C. Luong, D. Niyato, D. I. Kim and L. Wang, ``Efficient Training Management for Mobile Crowd-Machine Learning: A Deep Reinforcement Learning Approach,'' {\em IEEE Wireless Communications Letters}, vol. 8, no. 5, pp. 1345-1348, Oct. 2019.

\bibitem{zz3}
Y. Liu, G. Feng, Y. Sun, S. Qin and Y. -C. Liang, ``Device Association for RAN Slicing based on Hybrid Federated Deep Reinforcement Learning,'' {\em IEEE Transactions on Vehicular Technology}, pp. 1-15, 2020, doi: 10.1109/TVT.2020.3033035.

\bibitem{11}
F. Essa, K. Jambi, A. Fattouh, H. Al-Barhamtoshy, M. Khemakhem and A. Al-Ghamdi, ``A fedearted E-learning cloud system based on mixed reality," {\em IEEE/ACS 13th International Conference of Computer Systems and Applications (AICCSA), Agadir}, Nov.-Dec. 2016.

\bibitem{a1}
H. H. Yang, Z. Liu, T. Q. S. Quek and H. V. Poor, ``Scheduling Policies for Federated Learning in Wireless Networks," {\em IEEE Transactions on Communications}, vol. 68, no. 1, pp. 317-333, Jan. 2020.

\bibitem{12}
Y. Liu, J. Peng, J. Kang, A. M. Iliyasu, D. Niyato and A. A. A. El-Latif, ``A Secure Federated Learning Framework for 5G Networks," {\em IEEE Wireless Communications}, vol. 27, no. 4, pp. 24-31, Aug. 2020.

\bibitem{13}
H. Cha, J. Park, H. Kim, M. Bennis and S. L. Kim, ``Proxy Experience Replay: Federated Distillation for Distributed Reinforcement Learning," {\em IEEE Intelligent Systems}, vol. 35, no. 4, pp. 94-101, Jul.-Aug. 2020.

\bibitem{14}
J. Kang, Z. Xiong, D. Niyato, Y. Zou, Y. Zhang and M. Guizani, ``Reliable Federated Learning for Mobile Networks," {\em IEEE Wireless Communications}, vol. 27, no. 2, pp. 72-80, Apr. 2020.

\bibitem{15}
Y. Yu, K. Adu, N. Tashi, P. Anokye, X. Wang and M. A. Ayidzoe, ``RMAF: Relu-Memristor-Like Activation Function for Deep Learning,'' {\em IEEE Access}, vol. 8, pp. 72727-72741, Apr. 2020.

\bibitem{two2}
J. Chen, R. Monga, S. Bengio and R. Jozefowicz, ``Revisiting Distributed Synchronous SGD,'' {\em Computer Science}, 2016.

\end{thebibliography}
\end{document}